\RequirePackage{float}
\documentclass[a4paper,fleqn]{cas-dc}

\IfFileExists{stix.sty}{}{\usepackage{lmodern}}

\usepackage[numbers,sort&compress]{natbib}
\usepackage{amsmath,amssymb,amsfonts}
\usepackage{booktabs,multirow,graphicx,adjustbox,placeins,url}
\hypersetup{hidelinks}
\hypersetup{pdftitle={AnchorMixGAN: Anchor-Aligned Generative Semi-Supervision for DDoS Detection},pdfauthor={Jin Yang, Xufeng Liu, Yong Hu, Xueyang Wang, Honglu Yang, Gang~Li}}
\usepackage{tikz}
\usetikzlibrary{arrows.meta,positioning,fit,shapes.geometric}

\floatstyle{ruled}
\newfloat{algorithm}{tbp}{loa}
\floatname{algorithm}{Algorithm}

\ExplSyntaxOn
\RenewDocumentCommand \printorcid { } { }
\cs_set:Npn \__first_footerline: { }
\ExplSyntaxOff

\begin{document}
\raggedbottom
\shorttitle{}
\shortauthors{J. Yang et al.}

\title[mode=title]{AnchorMixGAN: Anchor-Aligned Generative Semi-Supervision for DDoS Detection in Cloud-Integrated IoT Networks}

\author[1,2,3]{Jin Yang}
\ead{yangjin66@scu.edu.cn}
\author[1]{Xufeng Liu}
\cormark[1]
\ead{liuxufeng@stu.scu.edu.cn}
\author[1]{Yong Hu}
\cormark[1]
\ead{huyong1946@outlook.com}
\author[1]{Xueyang Wang}
\author[4]{Honglu Yang}
\author[5]{Gang Li}
\cortext[1]{Corresponding authors.}

\affiliation[1]{organization={School of Cyber Science and Engineering, Sichuan University},city={Chengdu},postcode={610207},country={China}}
\affiliation[2]{organization={College of Information Science and Technology, Xizang University},city={Lhasa},postcode={850000},country={China}}
\affiliation[3]{organization={Shaanxi Key Laboratory of Intelligent Policing, Shaanxi Police College},city={Xi'an},postcode={710021},country={China}}
\affiliation[4]{organization={China Academy of Transportation Sciences, Ministry of Transport of the People's Republic of China},city={Beijing},postcode={100029},country={China}}
\affiliation[5]{organization={Kaiyuan Huachuang Technology (Group) Co., Ltd.},city={Beijing},postcode={102400},country={China}}

\begin{abstract}
Detecting distributed denial-of-service (DDoS) attacks in cloud-integrated IoT networks is difficult when labeled traffic is scarce. Generative semi-supervised learning can supplement the available training data, but prediction shifts induced by synthetic views may affect the targets assigned to real unlabeled flows. We propose AnchorMixGAN, a generative semi-supervised framework that addresses this problem through anchor-aligned target construction. Its Anchor-MAS module treats each real unlabeled flow as an anchor and creates alternative views by replacing one field group at a time with values from generated traffic. A frozen reference classifier predicts the anchor and its views; averaging and sharpening these predictions produces a soft target for the original flow. The flow and its target are then mixed with a labeled example using MixUp, allowing the detector to learn from both the original labeled records and the mixed examples. We analyze how reference-classifier error, view construction, and sharpening affect the target, and derive a bound on the resulting change in cross-entropy at a fixed detector prediction. At the reported 90\% training setting with 20\% of the training records labeled, AnchorMixGAN attains accuracies of 97.3\%, 97.4\%, and 96.5\% on NSL-KDD, BoT-IoT, and CICIoT2023, respectively, exceeding the corresponding MixGAN results by 1.6, 1.0, and 4.4 percentage points.
\end{abstract}
\begin{keywords}
Semi-supervised learning \sep Pseudo-labeling \sep Tabular data augmentation \sep DDoS detection
\end{keywords}
\maketitle

\section{Introduction}
\begin{figure}[pos=!htbp]
\centering
\includegraphics[width=\linewidth]{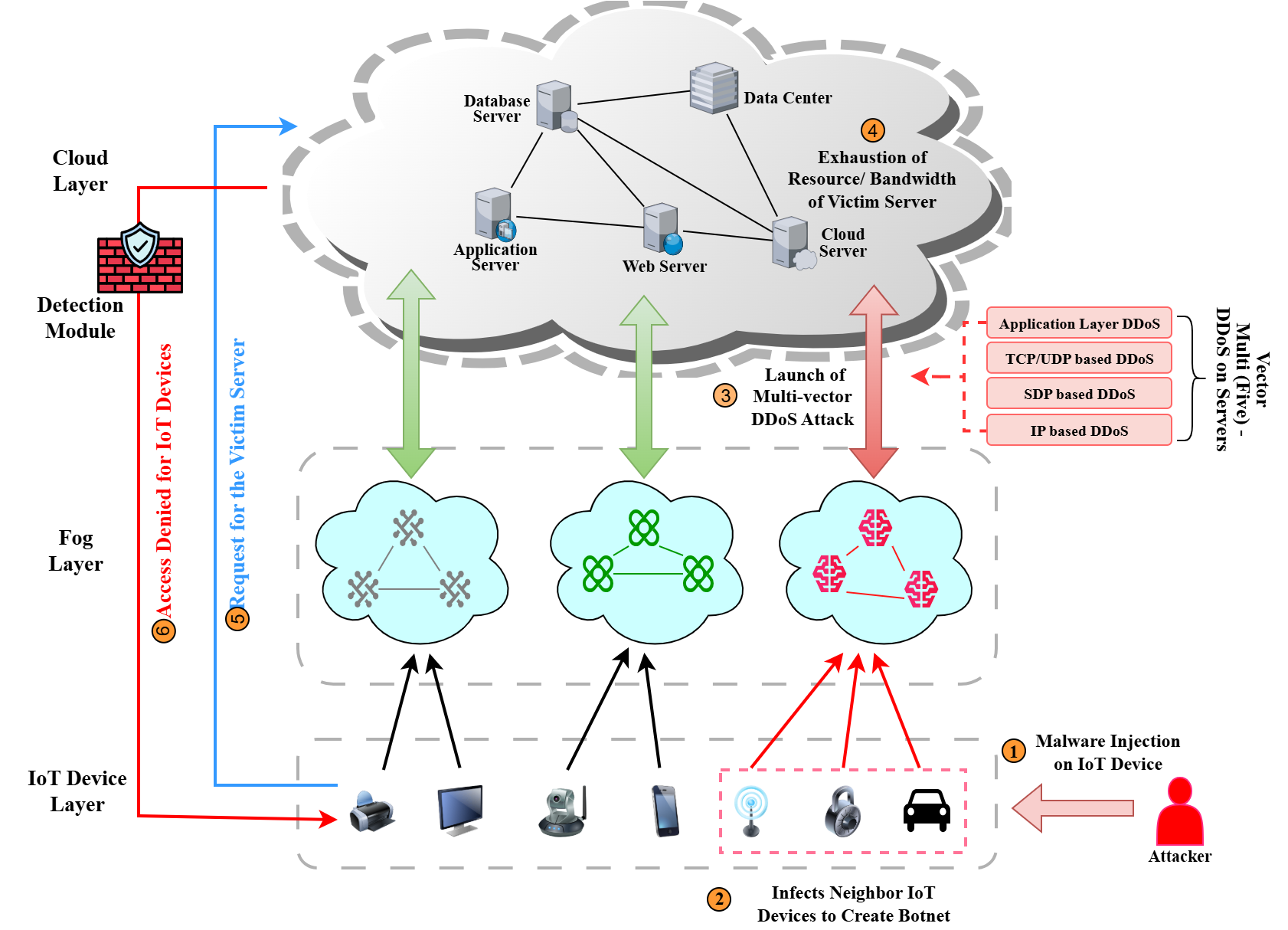}
\caption{DDoS threat scenario in a cloud-integrated IoT system.}
\label{fig:iot-scenario}
\end{figure}

Cloud services support the computing and storage needs of Internet of Things (IoT) devices, but they can also become targets of distributed denial-of-service (DDoS) attacks launched from compromised devices. Attack traffic consumes network bandwidth and server resources, interrupting services for legitimate users~\cite{agrawal2019defense,kumari2023comprehensive,bala2024ai}. Figure~\ref{fig:iot-scenario} illustrates this threat and the placement of the detection module. Flow records collected by gateways and network services provide the features used to identify attacks, yet obtaining reliable annotations for these records is costly. This motivates semi-supervised learning, which can use model-generated targets to incorporate unlabeled flows into detector training alongside labeled records. The central challenge is to construct reliable targets for these flows when only limited labeled data are available.

In view-based semi-supervised learning, targets can be estimated by aggregating predictions on alternative views of the same unlabeled record~\cite{berthelot2019mixmatch}. Generative models provide additional feature variations for view construction by synthesizing records from a learned traffic distribution~\cite{xu2019modeling}. MixGAN~\cite{wu2025mixgan} combines generative augmentation with semi-supervised target estimation, averaging and sharpening predictions on generated views to construct targets for unlabeled flows. However, the attribute changes introduced by these views may also alter the classifier's predictions, and these changes are subsequently incorporated into the target assigned to the original flow. When multiple views induce prediction shifts in the same direction, averaging may not eliminate the resulting deviation; sharpening increases the concentration of the prediction distribution but does not necessarily improve target accuracy. For tabular traffic containing categorical encodings and related numerical attributes, this issue calls for attention to where synthetic values are introduced, how field structure constrains their incorporation, and how the resulting views affect target estimation. MixGAN's theoretical analysis assumes pseudo-label accuracy without further separating the contributions of reference-classifier error, view construction, and sharpening to target error. Building on this framework, it is therefore necessary to specify the field-structure constraints governing generated views and characterize how view-induced prediction changes propagate to supervision targets and the training loss.

To this end, we propose AnchorMixGAN, which constructs soft targets through an anchor-aligned MixUp--Average--Sharpen (Anchor-MAS) module. The module treats each real unlabeled flow as an anchor and forms each donor view by replacing one complete field group with values from a single generated record while retaining all other fields. A frozen reference classifier then predicts the original flow and its donor views, and these predictions are averaged and sharpened to estimate a soft target. We analyze how this target construction affects supervision and evaluate AnchorMixGAN on three traffic benchmarks. Under the reported 90\% training-data setting with 20\% of the training records labeled, AnchorMixGAN achieves 96.5\% accuracy on CICIoT2023, exceeding MixGAN by 4.4 percentage points.

The main contributions are:
\begin{enumerate}
\item A target-construction method based on whole-group donor views. Anchor-MAS introduces generated values at the field-group level and combines predictions on the resulting views and the original flow to estimate an anchor-specific soft target.
\item A target-error decomposition that separates reference-classifier error, donor-view displacement, and sharpening displacement, together with a bound on their effect on cross-entropy at a fixed detector prediction.
\item An evaluation on three traffic datasets that compares AnchorMixGAN with MixGAN and representative detection methods, and examines performance under varying data and label availability, parameter sensitivity, and inference cost.
\end{enumerate}

The remainder of this paper is organized as follows. Section~2 reviews related work. Section~3 describes the detection task and the model foundations. Section~4 presents AnchorMixGAN and its analysis. Section~5 reports the experiments, and Section~6 concludes the paper.

\section{Related work}

\subsection{DDoS detection in cloud and IoT networks}
Learning-based DDoS detection addresses both traffic representation and deployment cost~\cite{agrawal2019defense,kumari2023comprehensive,bala2024ai,kumar2024comprehensive}. LUCID uses a lightweight convolutional detector~\cite{doriguzzi2020lucid}, while IoT and mobile-healthcare studies examine detection across heterogeneous traffic and multi-vector attacks~\cite{hizal2024novel,aguru2024lightweight}. Recurrent models and feature selection provide complementary approaches to attribute modeling and input reduction~\cite{syed2023fog,khanday2023implementation}.

Optimization-based approaches improve the learning procedure itself. FT-DBN combines deep belief networks with fuzzy-Taylor and optimization methods, and FACVO-DNFN uses an optimization-driven neuro-fuzzy model~\cite{velliangiri2020fuzzy,gsr2023facvo}. Related deep detectors also incorporate optimization~\cite{balasubramaniam2023optimization}. ADAM and SDN-based approaches connect classification decisions with attack mitigation~\cite{cai2023adam,hnamte2024ddos}.

Correlation-aware networks, FTG-Net-E, and TCG-IDS additionally model relationships among devices or observations~\cite{hekmati2024correlation,bakar2024ftg,wu2025tcg}. Limited labels remain a separate training concern~\cite{mvula2024survey}. Federated semi-supervised and active learning, confidence-based augmentation in Cycle-Fed, and hierarchical federated learning address distributed supervision and contaminated data~\cite{naeem2023federated,xiao2024cycle,gui2025solving}. Our study fixes the residual detector and examines supervision through target construction for individual flows.

\subsection{Generative augmentation for intrusion detection}
Augmentation supplies additional examples under class imbalance. SMOTE interpolates minority samples~\cite{chawla2002smote}, whereas CTGAN and CTAB-GAN learn tabular synthesis distributions~\cite{xu2019modeling,zhao2021ctab}. CTGAN-based botnet modeling, S2CGAN-IDS, and TMG-GAN use generated traffic with different representations and generation designs~\cite{habibi2023imbalanced,wang2023effective,ding2024tmg}.

Generation also supports detection objectives beyond sample expansion. GAN-AE combines adversarial learning and autoencoding; HDA-IDS connects signature and anomaly detection~\cite{boppana2023gan,li2024hda}. Continual AE-WGAN and SYN-GAN study streaming anomalies and synthetic-data classification, respectively~\cite{seghair2024continual,rahman2024syn}.

MixGAN incorporates generation into MAS averaging, sharpening, and interpolation~\cite{wu2025mixgan}. AnchorMixGAN develops target estimation at the field-group level: each generated record supplies a group of values for a real flow's view, whose prediction contributes to that flow's target.

\subsection{Semi-supervised targets and tabular views}
MixUp interpolates examples and labels~\cite{zhang2018mixup}; MixMatch averages augmented-input predictions and sharpens targets before interpolation~\cite{berthelot2019mixmatch}. Temporal Ensembling and Mean Teacher aggregate predictions or weights over training, while FixMatch supervises strong views with confident predictions~\cite{laine2017temporal,tarvainen2017mean,sohn2020fixmatch}.

For tabular inputs, VIME uses corruption and recovery, SubTab forms feature subsets, and SCARF applies corruption in contrastive learning~\cite{yoon2020vime,ucar2021subtab,bahri2022scarf}. Since feature choices affect intrusion detection~\cite{sarhan2024feature}, the view operator must specify the affected coordinates. Anchor-MAS combines complete-anchor predictions with predictions on predefined donor-group views, retaining the anchor--target pairing in MixUp.

Sharpening controls target concentration; probability quality concerns agreement with outcomes~\cite{guo2017calibration}. Preprocessing, label access, and model selection also affect the evaluation of semi-supervised detectors~\cite{oliver2018realistic,arp2022dos,kapoor2023leakage}.

\section{Detection task and model foundation}

\subsection{Notation and residual detector}
Let $\mathcal D_l=\{(x_i,y_i)\}_{i=1}^{N_l}$ contain labeled records and $\mathcal D_u=\{u_j\}_{j=1}^{N_u}$ contain records with hidden detector labels. Each preprocessed record belongs to $\mathbb R^d$, with $y\in\{0,1\}$ indicating non-DDoS and DDoS, respectively. A detector $f_\theta$ returns two logits, and $p_\theta(x)=\operatorname{softmax}(f_\theta(x))$ denotes its class probabilities. A separate supervised reference classifier has parameters $\theta_t$.

\begin{figure}[pos=!htbp]
\centering
\includegraphics[width=\linewidth]{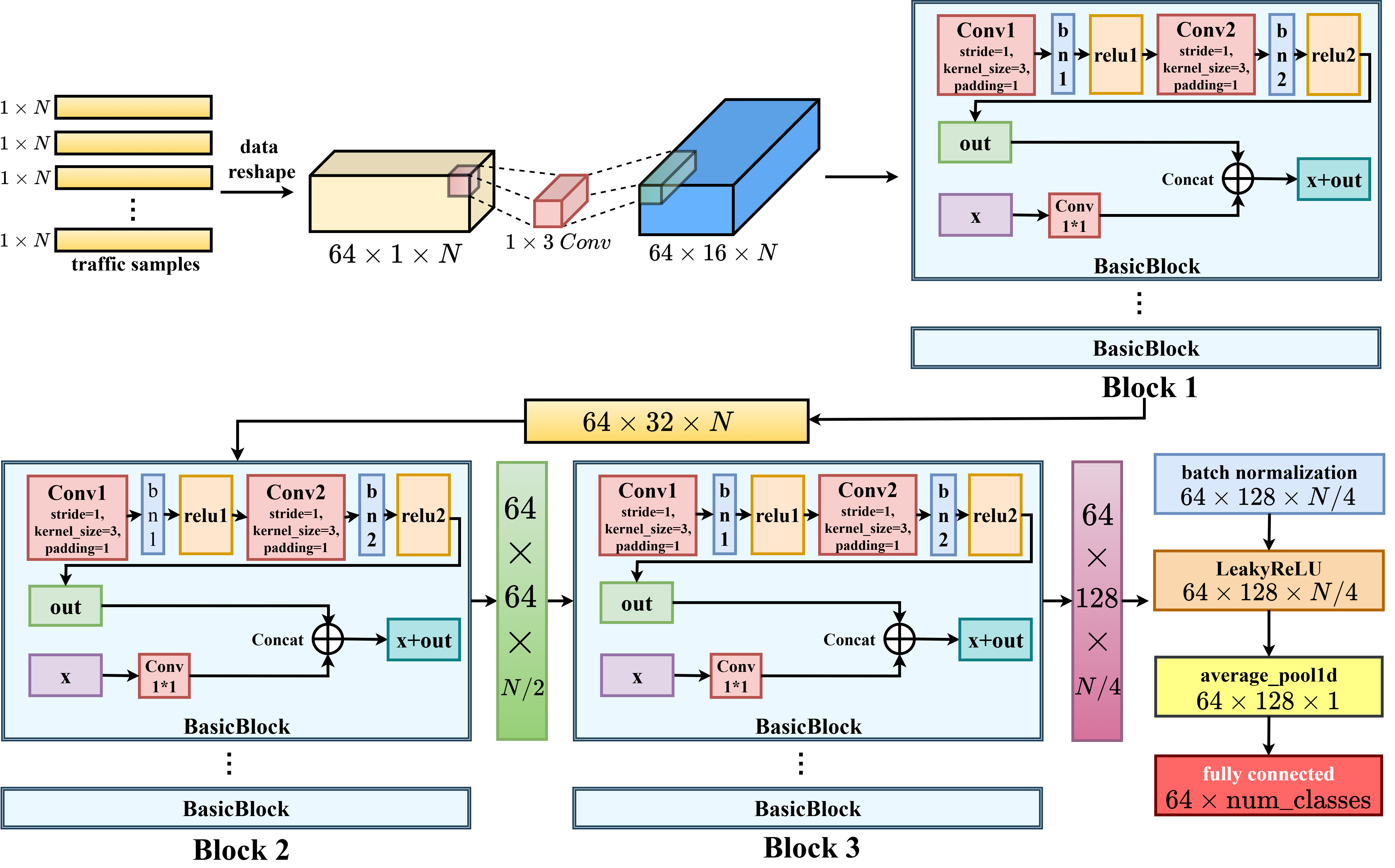}
\caption{Overall architecture of the 1-D WideResNet backbone.}
\label{fig:wideresnet}
\end{figure}

We use a one-dimensional WideResNet-28-2 backbone based on residual and wide-residual architectures~\cite{he2016deep,zagoruyko2016wide}. As illustrated in Figure~\ref{fig:wideresnet}, the input passes through an initial one-dimensional convolution and three residual stages with increasing channel widths. Batch normalization and LeakyReLU are used in feature extraction, followed by average pooling and a fully connected layer for binary classification. Each residual stage combines the transformed features with a shortcut connection:
\begin{equation}
z^{(r+1)}=S_r(z^{(r)})+F_r(z^{(r)};\theta_r),
\end{equation}
where $S_r$ is the identity or a projection matching dimensions. The convolutional axis follows the ordered feature vector of an individual flow, allowing the residual transformations to learn relationships among its attributes.

\subsection{MAS operations}
For a probability vector $a$, temperature sharpening is
\begin{equation}
\operatorname{Sharpen}(a,T)_c=\frac{a_c^{1/T}}{\sum_{r=0}^{1}a_r^{1/T}},\qquad T>0.
\label{eq:sharpen}
\end{equation}
For $T<1$, it increases the concentration of a nonuniform class distribution without changing its maximum-probability class. MAS uses an averaged target followed by this operation, then interpolates the unlabeled record and target with a labeled example. Anchor-MAS applies these operations to predictions on the complete anchor and its field-group views.

\section{AnchorMixGAN}
AnchorMixGAN integrates field-group view construction, anchor-dependent target estimation, and MixUp training in a generative semi-supervised detector.

\subsection{Overview}
For each real anchor, Anchor-MAS constructs views by substituting one field group at a time with generated donor values. Predictions on the complete anchor and its views are averaged to form the target. The sharpened probabilities are then used directly in MixUp. Training on both labeled records and mixed records provides direct label supervision as well as the soft targets constructed by the reference classifier.

\subsection{Grouped-field donor construction}
\begin{figure*}[pos=tbp]
\centering

\includegraphics[width=\textwidth]{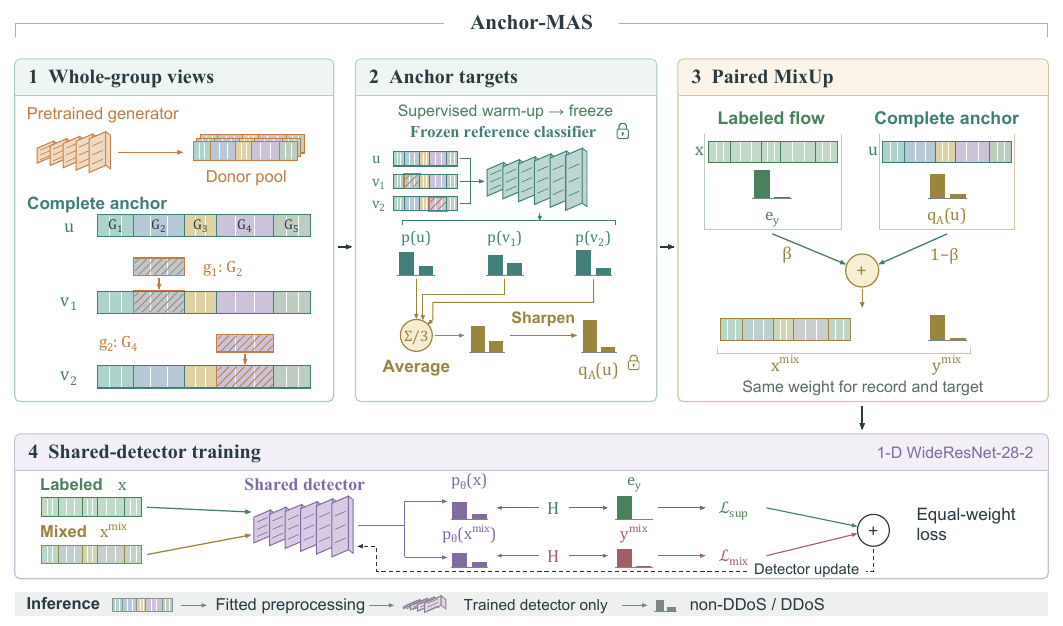}
\caption{Overview of AnchorMixGAN. (1) Generated donor groups replace complete groups of a real unlabeled anchor; unselected fields are retained. (2) One frozen reference classifier evaluates the complete anchor and two views. Its predictions are averaged and sharpened to form the cached soft target $q_A(u)$. (3) The complete anchor and its target are mixed with a labeled record and its label, using the same coefficient for features and targets. (4) One shared detector learns from labeled and mixed examples with $\mathcal L=\tfrac12\mathcal L_{sup}+\tfrac12\mathcal L_{mix}$; only detector parameters are updated. Repeated symbols across panels denote the same records and targets. Five field groups and the substituted positions are illustrative; $K=2$ views are shown. Reference classifier probabilities are abbreviated as $p(\cdot)$. Only fitted preprocessing and the trained detector are deployed.}
\label{fig:framework}
\end{figure*}

Partition the input coordinates into disjoint groups $\mathcal G=\{G_1,\ldots,G_m\}$. A group may represent a complete encoded categorical field or a designated set of related numerical attributes. For each real anchor $u$, draw donor records $g_1,\ldots,g_K$ from a retained synthetic pool and group masks $M_1,\ldots,M_K\in\{0,1\}^d$. Each mask selects all coordinates of one group. The corresponding view is
\begin{equation}
v_k(u,g_k)=(1-M_k)\odot u+M_k\odot g_k.
\label{eq:view}
\end{equation}
Unselected coordinates remain exactly equal to the anchor. All coordinates of a selected group come from the same donor. Figure~\ref{fig:framework} summarizes the training flow.

\subsection{Anchor-dependent pseudo-target}
First train $f_{\theta_t}$ using only $\mathcal D_l$ and freeze its parameters. For each anchor, combine its complete prediction with the predictions of its $K$ views:
\begin{align}
\bar q_A(u)&=\frac{p_{\theta_t}(u)+\sum_{k=1}^{K}p_{\theta_t}(v_k(u,g_k))}{K+1},\label{eq:anchor-average}\\
q_A(u)&=\operatorname{Sharpen}(\bar q_A(u),T).
\label{eq:anchor-target}
\end{align}
We compute these probabilities without gradients and cache the targets before detector training. The default settings are $K=2$ and $T=0.5$; the sensitivity study varies $T$. Donor choices, group masks, and resulting targets are fixed within each run.

\subsection{MixUp and detector optimization}
Pair an anchor $u_i$ with labeled record $(x_i,y_i)$, and draw $\beta_i\sim\operatorname{Beta}(\alpha,\alpha)$. Given the anchor-dependent target $q(u_i)=q_A(u_i)$,
\begin{align}
x_i^{mix}&=\beta_i x_i+(1-\beta_i)u_i,\label{eq:mix-features}\\
y_i^{mix}&=\beta_i e_{y_i}+(1-\beta_i)q(u_i),
\label{eq:mix}
\end{align}
where $e_{y_i}$ is the one-hot class label. The default is $\alpha=0.75$. The sampled coefficient is used directly in both interpolations. Donor views determine the pseudo-target; the complete real anchor is the input interpolated in Eq.~\eqref{eq:mix-features}.

Let $H(a,b)=-\sum_c a_c\log b_c$. The detector minimizes the average cross-entropy over equal numbers of genuine labeled and interpolated examples:
\begin{align}
\mathcal L_{sup}&=\frac{1}{N_l}\sum_i H(e_{y_i},p_\theta(x_i)),\\
\mathcal L_{mix}&=\frac{1}{N_l}\sum_i H(y_i^{mix},p_\theta(x_i^{mix})),\\
\mathcal L&=\tfrac12\mathcal L_{sup}+\tfrac12\mathcal L_{mix}.
\label{eq:loss}
\end{align}
The supervised term trains the detector on the original labels, and the mixed term uses the full soft-label distributions. We initialize the detector randomly and update only its parameters with Eq.~\eqref{eq:loss}. The reference classifier remains frozen. Algorithm~\ref{alg:train} details the complete procedure.

\begin{algorithm}[tbp]
\caption{AnchorMixGAN training}
\label{alg:train}
\small
\textbf{Input:} Preprocessed labeled set $\mathcal D_l$, unlabeled set $\mathcal D_u$, donor pool $\mathcal P$, field groups $\mathcal G$;
parameters $K,T,\alpha$ and detector update budget $S$.\\
\textbf{Output:} Trained detector $f_{\theta^\star}$.
\par\smallskip
\begingroup
\setlength{\tabcolsep}{0pt}
\renewcommand{\arraystretch}{1.12}
\begin{tabular}{@{}p{\linewidth}@{}}
\textit{// Step 1: Construct anchor-dependent targets}\\
Train $f_{\theta_t}$ on $\mathcal D_l$; freeze in evaluation mode.\\
Select $N_l=|\mathcal D_l|$ real anchors and fix the pairs $(x_i,u_i)$.\\
\textbf{for} $i=1,\ldots,N_l$ \textbf{do}\\
\hspace*{1em}Sample $K$ donors $g_{ik}\in\mathcal P$ and whole-group masks $M_{ik}$.\\
\hspace*{1em}$v_{ik}\leftarrow(1-M_{ik})\odot u_i+M_{ik}\odot g_{ik}$.\\
\hspace*{1em}$\bar q_i\leftarrow\big[p_{\theta_t}(u_i)+\sum_{k=1}^{K}p_{\theta_t}(v_{ik})\big]/(K+1)$.\\
\hspace*{1em}Cache $q_i\leftarrow\operatorname{stopgrad}[\operatorname{Sharpen}(\bar q_i,T)]$.\\
\textbf{end for}\\[2pt]
\textit{// Step 2: Apply paired MixUp once}\\
\textbf{for} $i=1,\ldots,N_l$ \textbf{do}\\
\hspace*{1em}Sample $\beta_i\sim\operatorname{Beta}(\alpha,\alpha)$.\\
\hspace*{1em}$x_i^{mix}\leftarrow\beta_i x_i+(1-\beta_i)u_i$.\\
\hspace*{1em}$y_i^{mix}\leftarrow\beta_i e_{y_i}+(1-\beta_i)q_i$.\\
\textbf{end for}\\
Form $\mathcal S$ with $N_l$ genuine pairs $(x_i,e_{y_i})$ and $N_l$ mixed pairs $(x_i^{mix},y_i^{mix})$.\\[2pt]
\textit{// Step 3: Update the shared detector}\\
Randomly initialize $\theta$ and its Adam optimizer.\\
\textbf{for} $s=1,\ldots,S$ \textbf{do}\\
\hspace*{1em}Take the next batch $B$ from shuffled passes through the fixed list $\mathcal S$.\\
\hspace*{1em}$\widehat{\mathcal L}_B\leftarrow |B|^{-1}\sum_{(z,t)\in B}H(t,p_\theta(z))$.\\
\hspace*{1em}Update only $\theta$ with Adam.\\
\textbf{end for}\\
\textbf{return} the final detector $f_{\theta^\star}$.\\
\end{tabular}
\endgroup
\end{algorithm}

\subsection{Anchor dependence and target displacement}
Equation~\eqref{eq:anchor-average} averages predictions on $u$ and views that retain its unselected fields. For a fixed reference classifier, subtracting the complete-anchor prediction from this average gives the following bound:
\begin{equation}
\begin{aligned}
\left\|\bar q_A(u)-p_{\theta_t}(u)\right\|_1 &\leq D(u),\\
D(u)&:=\frac{1}{K+1}\sum_{k=1}^{K}
\left\|p_{\theta_t}(v_k)-p_{\theta_t}(u)\right\|_1.
\end{aligned}
\label{eq:deviation}
\end{equation}
The triangle inequality yields Eq.~\eqref{eq:deviation}. Each view contributes its prediction difference from the complete anchor with weight $1/(K+1)$, so $D(u)$ bounds the displacement introduced by averaging.

If the reference classifier probability map is $L_p$-Lipschitz on the relevant inputs, then
\begin{equation}
D(u)\leq\frac{L_p}{K+1}\sum_{k=1}^{K}
\left\|M_k\odot(g_k-u)\right\|_2.
\label{eq:lipschitz}
\end{equation}
The mask restricts the input difference to the substituted field group. This bound relates target displacement to the replacement magnitude and the local sensitivity of the reference classifier.

\subsection{Reference classifier error, view sensitivity, and sharpening}
Let $r(u)$ denote the true conditional class distribution of the real anchor. Define the reference classifier error $e_t(u)=\|p_{\theta_t}(u)-r(u)\|_1$ and the sharpening displacement $S_T(u)=\|q_A(u)-\bar q_A(u)\|_1$. Applying the triangle inequality around the unsharpened average and the complete-anchor prediction gives
\begin{equation}
\|q_A(u)-r(u)\|_1\leq e_t(u)+D(u)+S_T(u).
\label{eq:target-error}
\end{equation}
The three terms separate reference classifier error on the complete anchor, donor-view displacement, and sharpening displacement. When $T=1$, $S_T(u)=0$; for $T<1$, sharpening concentrates the distribution, and $S_T(u)$ measures the resulting change in the target.

We next bound the effect of target error on cross-entropy. For a fixed detector probability vector $s$ with $s_c\geq\varepsilon>0$ for both classes,
\begin{align}
&|H(q_A(u),s)-H(r(u),s)|\nonumber\\
&\quad\leq\log(1/\varepsilon)\,\|q_A(u)-r(u)\|_1\\
&\quad\leq\log(1/\varepsilon)\,[e_t(u)+D(u)+S_T(u)].
\label{eq:target-loss}
\end{align}
Expanding the cross-entropy difference and using $|\log s_c|\leq\log(1/\varepsilon)$ gives the first inequality; substituting Eq.~\eqref{eq:target-error} gives the second. Equation~\eqref{eq:target-loss} bounds the change in cross-entropy caused by target displacement at a fixed detector prediction $s$.

\subsection{Computation and inference}
Before detector training, Anchor-MAS makes $K+1$ reference classifier evaluations per selected anchor and caches the result. Constructing views costs $O(KN_a d)$ field operations, where $N_a$ is the number of anchors, in addition to obtaining donors. At inference, fitted preprocessing and one trained detector process each flow. Section~\ref{subsec:efficiency} reports detector-forward timings under matched hardware and batch-size settings.

\section{Experiments}
\label{sec:experiment}

\subsection{Datasets}

To evaluate the performance of the proposed AnchorMixGAN framework across diverse network conditions, we utilize three publicly available benchmark datasets: NSL-KDD~\cite{tavallaee2009detailed}, BoT-IoT~\cite{koroniotis2019towards}, and CICIoT2023~\cite{neto2023ciciot2023}.

NSL-KDD is an improved version of the original KDD'99 dataset, covering DoS, Probe, R2L, and normal traffic. To address the shortcomings of the original dataset, the NSL-KDD dataset removes redundant records and performs reasonable sampling of attack types, making the evaluation results more reliable.

BoT-IoT is generated through an IoT testing platform built in a virtualized environment, combining normal traffic from various IoT devices with different types of malicious botnet traffic. The attack types include DDoS, DoS, port scanning, information theft, and more. Compared with many existing datasets, BoT-IoT provides greater diversity in attack scenarios and more detailed traffic labels, making it suitable for network forensics analysis and IDS evaluation.

CICIoT2023 is collected by the CIC IoT Lab, which deploys an extensive topology consisting of 105 real IoT devices. It covers seven attack categories, including DDoS, DoS, reconnaissance attacks, web attacks, brute-force attacks, spoofing attacks, and Mirai. The dataset provides fine-grained labels and is used for evaluating detection in complex IoT-cloud environments with diverse protocols.

\subsection{Evaluation Metrics}
To evaluate the final performance of the trained model, we employed five classic metrics: accuracy, precision, recall, F1-score, and macro-averaged TPR/TNR. These metrics collectively offer a comprehensive validation of the model's performance, aligning with standard practices for evaluating DDoS detection methods. Their calculation formulas are presented below, where TP, TN, FP, and FN denote the number of true positives, true negatives, false positives, and false negatives, respectively, and C represents the number of classes.
\begingroup
\allowdisplaybreaks[4]
\begin{align}
\mathrm{Accuracy}      & = \mathrm{\frac{TP+TN}{TP+TN+FP+FN}} \label{eq:acc}\\
\mathrm{Precision}     & = \mathrm{\frac{TP}{TP+FP}}           \label{eq:prec}\\
\mathrm{Recall}        & = \mathrm{\frac{TP}{TP+FN}}           \label{eq:rec}\\
\mathrm{F1\text{-score}}     & = \mathrm{\frac{2TP}{2TP+FP+FN}}      \label{eq:f1}\\
\mathrm{TPR}_{\text{macro}} &= \frac{1}{C}\sum_{i=1}^{C}\mathrm{\frac{TP_i}{TP_i+FN_i}} \label{eq:tprmacro}\\
\mathrm{TNR}_{\text{macro}} &= \frac{1}{C}\sum_{i=1}^{C}\mathrm{\frac{TN_i}{TN_i+FP_i}}. \label{eq:tnrmacro}
\end{align}
\endgroup

\subsection{Experimental Setup}
The data preparation and training settings follow the conference protocol~\cite{wu2025mixgan}. The sampled datasets contain 100,000 NSL-KDD records, 200,000 BoT-IoT records, and 300,000 CICIoT2023 records. The reference training sets contain 94,000/2,000, 170,000/10,000, and 260,000/20,000 non-DDoS/DDoS records, respectively; the corresponding test sets contain 2,000 records per class for NSL-KDD and 10,000 records per class for BoT-IoT and CICIoT2023.

The training-data ratio specifies the fraction of records used for training in the corresponding experiment. The labeling ratio specifies the fraction of that training set whose labels are available to the semi-supervised detector. The default labeling ratio is 20\%, with the remaining 80\% treated as unlabeled. Table~\ref{tab:nsl_bot_comparison} varies the training-data ratio from 60\% to 90\%. Tables~\ref{tab:conference-cic}, \ref{tab:conference-ablation}, \ref{tab:mas_hyperparams}, and~\ref{tab:label_ratio_bot} use the 90\% training setting. The limited-label experiment varies the labeling ratio from 1\% to 20\% while retaining that training-data ratio.

\subsection{Implementation Details}

\begin{table*}[pos=htbp]
\centering
\caption{Complete Performance Metrics Comparison on NSL-KDD and BoT-IoT Datasets}
\label{tab:nsl_bot_comparison}
\scriptsize
\begin{adjustbox}{max width=\textwidth}
\begin{tabular}{ll|cc|cc|cc|cc|cc|cc|cc|cc}
\toprule
\multirow{2}{*}{Metric} & \multirow{2}{*}{Train Ratio} 
& \multicolumn{2}{c|}{LUCID}
& \multicolumn{2}{c|}{FT-DBN}
& \multicolumn{2}{c|}{HLBO+DSA}
& \multicolumn{2}{c|}{FACVO-DNFN}
& \multicolumn{2}{c|}{GHLBO+DSA}
& \multicolumn{2}{c|}{STKD}
& \multicolumn{2}{c|}{MixGAN}
& \multicolumn{2}{c}{AnchorMixGAN} \\
\cmidrule(lr){3-4} \cmidrule(lr){5-6} \cmidrule(lr){7-8} \cmidrule(lr){9-10} \cmidrule(lr){11-12} \cmidrule(lr){13-14} \cmidrule(lr){15-16} \cmidrule(lr){17-18}
& & NSL & BoT & NSL & BoT & NSL & BoT & NSL & BoT & NSL & BoT & NSL & BoT & NSL & BoT & NSL & BoT\\
\midrule
\textbf{Accuracy} 
& 60\% & 0.806 & 0.797 & 0.884 & 0.863 & 0.857 & 0.885 & 0.904 & 0.887 & 0.899 & 0.903 & 0.920 & 0.906 & 0.923 & 0.916 & \textbf{0.947} & \textbf{0.953} \\
& 70\% & 0.813 & 0.803 & 0.890 & 0.867 & 0.843 & 0.891 & 0.911 & 0.891 & 0.902 & 0.908 & 0.927 & 0.919 & 0.931 & 0.929 & \textbf{0.950} & \textbf{0.957} \\
& 80\% & 0.835 & 0.835 & 0.902 & 0.880 & 0.891 & 0.895 & 0.921 & 0.902 & 0.908 & 0.912 & 0.939 & 0.926 & 0.943 & 0.951 & \textbf{0.968} & \textbf{0.967} \\
& 90\% & 0.847 & 0.841 & 0.909 & 0.890 & 0.897 & 0.899 & 0.930 & 0.914 & 0.914 & 0.917 & 0.941 & 0.933 & 0.957 & 0.964 & \textbf{0.973} &  \textbf{0.974} \\
\midrule
\textbf{Precision} 
& 60\% & 0.781 & 0.820 & 0.831 & 0.812 & 0.877 & 0.870 & 0.850 & 0.834 & 0.862 & 0.878 & 0.913 & 0.892 & 0.914 & 0.916 & \textbf{0.950} & \textbf{0.954} \\
& 70\% & 0.801 & 0.826 & 0.837 & 0.815 & 0.865 & 0.876 & 0.856 & 0.837 & 0.871 & 0.885 & 0.918 & 0.901 & 0.920 & 0.933 & \textbf{0.946} & \textbf{0.956} \\
& 80\% & 0.815 & 0.834 & 0.848 & 0.827 & 0.882 & 0.883 & 0.866 & 0.848 & 0.883 & 0.890 & 0.926 & 0.916 & 0.927 & 0.953 & \textbf{0.969} & \textbf{0.966} \\
& 90\% & 0.826 & 0.837 & 0.855 & 0.837 & 0.879 & 0.887 & 0.875 & 0.859 & 0.902 & 0.908 & 0.934 & 0.922 & 0.944 & 0.963 & \textbf{0.969} &  \textbf{0.974} \\
\midrule
\textbf{Recall} 
& 60\% & 0.819 & 0.826 & 0.849 & 0.828 & 0.856 & 0.879 & 0.849 & 0.828 & 0.895 & 0.884 & 0.905 & 0.889 & 0.914 & 0.916 & \textbf{0.943} & \textbf{0.952} \\
& 70\% & 0.826 & 0.827 & 0.865 & 0.865 & 0.862 & 0.883 & 0.865 & 0.865 & 0.895 & 0.891 & 0.917 & 0.902 & 0.926 & 0.929 & \textbf{0.954} & \textbf{0.957} \\
& 80\% & 0.832 & 0.832 & 0.883 & 0.879 & 0.865 & 0.889 & 0.883 & 0.879 & 0.907 & 0.896 & 0.928 & 0.916 & 0.953 & 0.951 & \textbf{0.966} & \textbf{0.968} \\
& 90\% & 0.842 & 0.838 & 0.909 & 0.889 & 0.868 & 0.894 & 0.909 & 0.889 & 0.909 & 0.908 & 0.937 & 0.921 & 0.964 & 0.965 & \textbf{0.976} &  \textbf{0.973} \\
\midrule
\textbf{F1-Score} 
& 60\% & 0.800 & 0.823 & 0.829 & 0.808 & 0.861 & 0.874 & 0.849 & 0.831 & 0.878 & 0.881 & 0.910 & 0.890 & 0.914 & 0.916 & \textbf{0.946} & \textbf{0.953} \\
& 70\% & 0.813 & 0.827 & 0.839 & 0.819 & 0.860 & 0.879 & 0.861 & 0.851 & 0.883 & 0.888 & 0.917 & 0.901 & 0.923 & 0.929 & \textbf{0.950} & \textbf{0.957} \\
& 80\% & 0.823 & 0.833 & 0.855 & 0.837 & 0.873 & 0.886 & 0.874 & 0.863 & 0.895 & 0.893 & 0.927 & 0.916 & 0.938 & 0.950 & \textbf{0.967} &  \textbf{0.967} \\
& 90\% & 0.834 & 0.838 & 0.864 & 0.847 & 0.874 & 0.890 & 0.891 & 0.874 & 0.902 & 0.908 & 0.936 & 0.922 & 0.952 & 0.963 & \textbf{0.973} &  \textbf{0.973} \\
\midrule
\textbf{TPR} 
& 60\% & 0.819 & 0.826 & 0.826 & 0.804 & 0.856 & 0.879 & 0.849 & 0.828 & 0.895 & 0.865 & 0.904 & 0.889 & 0.914 & 0.916 & \textbf{0.943} & \textbf{0.952} \\
& 70\% & 0.826 & 0.827 & 0.840 & 0.823 & 0.862 & 0.883 & 0.865 & 0.865 & 0.895 & 0.872 & 0.917 & 0.902 & 0.926 & 0.929 & \textbf{0.954} & \textbf{0.957} \\
& 80\% & 0.832 & 0.832 & 0.863 & 0.846 & 0.865 & 0.889 & 0.883 & 0.879 & 0.907 & 0.878 & 0.928 & 0.924 & 0.953 & 0.951 & \textbf{0.966} & \textbf{0.968} \\
& 90\% & 0.842 & 0.838 & 0.874 & 0.858 & 0.868 & 0.894 & 0.909 & 0.889 & 0.909 & 0.883 & 0.937 & 0.938 & 0.964 & 0.963 & \textbf{0.976} & \textbf{0.973} \\
\midrule
\textbf{TNR} 
& 60\% & 0.771 & 0.819 & 0.857 & 0.832 & 0.879 & 0.884 & 0.893 & 0.854 & 0.857 & 0.872 & 0.878 & 0.877 & 0.914 & 0.916 & \textbf{0.950} & \textbf{0.954} \\
& 70\% & 0.795 & 0.826 & 0.868 & 0.845 & 0.872 & 0.891 & 0.897 & 0.869 & 0.867 & 0.878 & 0.883 & 0.881 & 0.920 & 0.933 & \textbf{0.946} & \textbf{0.956} \\
& 80\% & 0.811 & 0.834 & 0.878 & 0.854 & 0.884 & 0.896 & 0.900 & 0.889 & 0.880 & 0.884 & 0.895 & 0.917 & 0.927 & 0.953 & \textbf{0.969} & \textbf{0.966} \\
& 90\% & 0.823 & 0.837 & 0.889 & 0.875 & 0.881 & 0.908 & 0.929 & 0.900 & 0.901 & 0.909 & 0.899 & 0.918 & 0.944 & 0.954 & \textbf{0.969} & \textbf{0.974} \\
\bottomrule
\end{tabular}
\end{adjustbox}
\end{table*}
\textit{Preprocessing.} Constant-valued features are removed, missing values are imputed by the mean, and categorical attributes are one-hot encoded where applicable. Features are standardized using z-score normalization. Random Forest feature importance is used to retain the top 15 features, with a 70/30 training--validation split for feature ranking. The selected feature representation is then held fixed across the training-ratio experiments for each dataset.

\textit{Generation and view construction.} CTGAN supplies class-conditional synthetic traffic. Its generator takes a 128-dimensional noise vector and a class condition; the generator and discriminator use three-layer multilayer perceptrons with LeakyReLU activations. The generator is trained offline for 500 epochs using a Wasserstein objective with gradient penalty coefficient 10. Generated records are placed in the same feature representation as the real flows and retained as a donor pool. Anchor-MAS draws two donors per anchor and replaces one complete field group in each view. Encoded coordinates of a categorical field are kept together, and related numerical attributes are assigned to predefined groups. The donor choices and group masks remain fixed during detector training.

\textit{Detector training.} AnchorMixGAN is implemented in PyTorch with a 1-D WideResNet-28-2 detector. The reference classifier is trained on the available labeled records and then frozen for target construction. We train the detector for 10 epochs using Adam~\cite{kingma2015adam}, a learning rate of $10^{-3}$, and a batch size of 64. The default Anchor-MAS parameters are $K=2$, $\alpha=0.75$, and $T=0.5$. Training uses the original labeled records and the soft-labeled mixed examples defined in Section~4.4.

\subsection{Comparison with Representative Methods}

To evaluate AnchorMixGAN, we compare it with LUCID~\cite{doriguzzi2020lucid}, FT-DBN~\cite{velliangiri2020fuzzy}, HLBO+DSA~\cite{dehghani2022hybrid,balasubramaniam2023optimization}, FACVO-DNFN~\cite{gsr2023facvo}, GHLBO+DSA~\cite{balasubramaniam2023optimization}, and STKD~\cite{wang2024spatial} on NSL-KDD and BoT-IoT. These baselines cover convolutional detection, deep belief networks, optimization-based deep learning, neuro-fuzzy detection, and knowledge distillation. MixGAN provides an additional baseline across all three datasets.

\begin{figure*}[pos=!t]
\centering
\includegraphics[width=\textwidth]{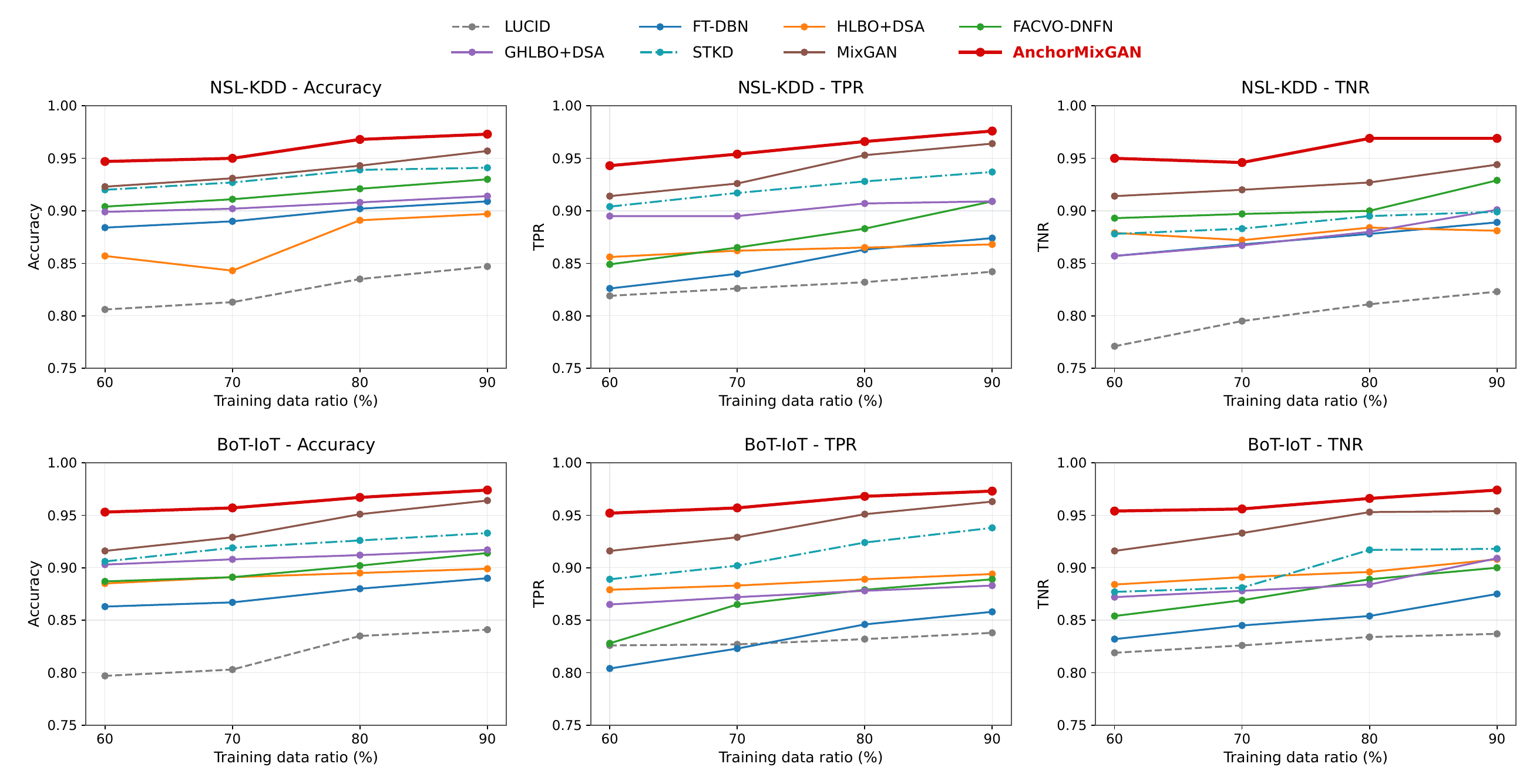}
\caption{Performance comparison on NSL-KDD (top row) and BoT-IoT (bottom row) at training data ratios of 60\%--90\%. Columns show accuracy, TPR, and TNR, respectively. Curves correspond to the values in Table~\ref{tab:nsl_bot_comparison}, with AnchorMixGAN highlighted in red. All panels use the same vertical scale.}
\label{fig:nsl-bot-comparison}
\end{figure*}

We examine overall detection performance and its dependence on training-data availability. Table~\ref{tab:nsl_bot_comparison} reports six metrics on NSL-KDD and BoT-IoT at training ratios of 60\%, 70\%, 80\%, and 90\%, while Figure~\ref{fig:nsl-bot-comparison} shows the corresponding accuracy, TPR, and TNR trends. Table~\ref{tab:conference-cic} extends the comparison to CICIoT2023 at the 90\% training setting.

\begin{table}[pos=!htbp]
\centering\footnotesize
\caption{Method comparison on CICIoT2023 at the 90\% training setting.}
\label{tab:conference-cic}
\setlength{\tabcolsep}{2.5pt}
\renewcommand{\arraystretch}{1.08}
\begin{adjustbox}{max width=\linewidth}
\begin{tabular}{lrrrrrr}
\toprule
Method & Acc. & Prec. & Recall & F1 & TPR & TNR\\
\midrule
Simple RNN & 0.903 & 0.880 & 0.932 & 0.905 & 0.932 & 0.873\\
BiLSTM & 0.878 & 0.899 & 0.850 & 0.874 & 0.849 & 0.905\\
CLGAN & 0.909 & 0.923 & 0.909 & 0.908 & 0.909 & 0.923\\
STACKING & 0.919 & 0.877 & \textbf{0.975} & 0.924 & \textbf{0.975} & 0.864\\
MixGAN & 0.921 & 0.925 & 0.921 & 0.920 & 0.921 & 0.925\\
AnchorMixGAN & \textbf{0.965} & \textbf{0.972} & 0.958 & \textbf{0.965} & 0.958 & \textbf{0.972}\\

\bottomrule
\end{tabular}
\end{adjustbox}
\end{table}

AnchorMixGAN achieves the highest accuracy among the compared methods on all three datasets. At the 90\% training setting, it reaches 97.3\% accuracy on NSL-KDD and 97.4\% on BoT-IoT, improving on MixGAN by 1.6 and 1.0 percentage points, respectively (Table~\ref{tab:nsl_bot_comparison}). These gains are accompanied by improvements in both TPR and TNR. On NSL-KDD, TPR rises from 96.4\% to 97.6\%, while TNR rises from 94.4\% to 96.9\%. On BoT-IoT, the corresponding changes are 96.3\% to 97.3\% and 95.4\% to 97.4\%. Thus, AnchorMixGAN improves attack detection while reducing false alarms on non-DDoS traffic.

The largest accuracy gain over MixGAN occurs on CICIoT2023 (Table~\ref{tab:conference-cic}). AnchorMixGAN achieves 96.5\% accuracy and 96.5\% F1-score, exceeding MixGAN by 4.4 and 4.5 percentage points, respectively. It also leads the comparison in precision and TNR, both reaching 97.2\%. The comparison with STACKING illustrates the different detection profiles: STACKING has a higher TPR (97.5\% versus 95.8\%), whereas AnchorMixGAN raises TNR from 86.4\% to 97.2\% and accuracy from 91.9\% to 96.5\%. AnchorMixGAN therefore combines high attack detection with substantially fewer false positives, yielding the best accuracy and F1-score in this comparison.

Figure~\ref{fig:nsl-bot-comparison} shows that AnchorMixGAN maintains its lead in accuracy, TPR, and TNR across all four training ratios on both datasets. Accuracy rises from 94.7\% to 97.3\% on NSL-KDD and from 95.3\% to 97.4\% on BoT-IoT. The gains over MixGAN range from 1.6 to 2.5 percentage points on NSL-KDD and from 1.0 to 3.7 percentage points on BoT-IoT. The BoT-IoT accuracy curves are furthest apart at the 60\% training ratio and gradually converge as more data become available, showing a particularly strong advantage for AnchorMixGAN in the smaller-data configurations. The TPR and TNR panels further show that the improvement covers both attack detection and non-DDoS recognition throughout the evaluated range.

\subsection{Component Analysis}
To analyze the contribution of each component to detection performance, Table~\ref{tab:conference-ablation} compares the base network, models with data augmentation or semi-supervised training, and the complete AnchorMixGAN framework at the 90\% training setting. Base uses the supervised backbone alone. Base+SMOTE and Base+CTGAN add data augmentation, Base+MAS introduces semi-supervised training, and MixGAN combines generation with MAS. AnchorMixGAN uses anchor-aligned target construction and trains on both labeled and soft-labeled mixed records.

\begin{table}[pos=!htbp]
\centering\small
\caption{Component comparison at the 90\% training setting. Reference configurations (Base through MixGAN) are from~\cite{wu2025mixgan}.}
\label{tab:conference-ablation}
\setlength{\tabcolsep}{5pt}
\renewcommand{\arraystretch}{1.08}
\begin{adjustbox}{max width=\linewidth}
\begin{tabular}{lrrrrrrrrr}
\toprule
\multirow{2}{*}{Configuration} & \multicolumn{3}{c}{NSL-KDD} & \multicolumn{3}{c}{BoT-IoT} & \multicolumn{3}{c}{CICIoT2023}\\
\cmidrule(lr){2-4}\cmidrule(lr){5-7}\cmidrule(lr){8-10}
 & Acc. & TPR & TNR & Acc. & TPR & TNR & Acc. & TPR & TNR\\
\midrule
Base & 0.903 & 0.879 & 0.931 & 0.891 & 0.891 & 0.911 & 0.810 & 0.809 & 0.826\\
Base+SMOTE & 0.911 & 0.868 & 0.940 & 0.902 & 0.901 & 0.914 & 0.852 & 0.851 & 0.858\\
Base+CTGAN & 0.931 & 0.901 & 0.953 & 0.939 & 0.931 & 0.938 & 0.906 & 0.905 & 0.907\\
Base+MAS & 0.922 & 0.884 & 0.943 & 0.925 & 0.924 & 0.925 & 0.892 & 0.891 & 0.891\\
MixGAN & 0.957 & 0.964 & 0.944 & 0.965 & 0.965 & 0.954 & 0.921 & 0.920 & 0.925\\
AnchorMixGAN & \textbf{0.973} & \textbf{0.976} & \textbf{0.969} &  \textbf{0.974} & \textbf{0.973} & \textbf{0.974} & \textbf{0.965} & \textbf{0.958} & \textbf{0.972}\\

\bottomrule
\end{tabular}
\end{adjustbox}
\end{table}

The component comparison shows a clear benefit from combining generation and semi-supervised training. On BoT-IoT, accuracy increases from 89.1\% for Base to 93.9\% with CTGAN and 92.5\% with MAS, while their combination in MixGAN reaches 96.5\%. AnchorMixGAN further increases accuracy to 97.4\%. AnchorMixGAN achieves the highest accuracy, TPR, and TNR in the component comparison across all three datasets. It also achieves the highest accuracy in the full-method comparisons. Relative to MixGAN in Table~\ref{tab:conference-ablation}, the accuracy gains are 1.6 percentage points on NSL-KDD, 0.9 on BoT-IoT, and 4.4 on CICIoT2023.

We next examine how the mixing and sharpening parameters affect detection accuracy. Table~\ref{tab:mas_hyperparams} reports BoT-IoT results at the 90\% training setting. With $T=0.7$, increasing $\alpha$ from 0.40 to 0.75 raises accuracy from 96.16\% to 96.93\%, whereas a further increase to 0.90 reduces it to 95.82\%. At $\alpha=0.75$, $T=0.5$ achieves 97.29\%, outperforming both $T=0.3$ (96.85\%) and $T=0.7$ (96.93\%). Accuracy remains above 95.8\% throughout the tested combinations, with a total spread of 1.47 percentage points. These results favor $\alpha=0.75$ and $T=0.5$ and show that neither a larger mixing parameter nor stronger sharpening necessarily improves detection.

\begin{table}[pos=!ht]
\centering
\caption{Impact of $\alpha$ and $T$ on BoT-IoT Detection Performance (90\% Training Data)}
\scriptsize
\label{tab:mas_hyperparams}
\begin{tabular}{ccc}
\toprule
$\alpha$ & $T$ & Accuracy (\%) \\
\midrule
0.40 & 0.7 & 96.16 \\
0.75 & 0.7 & 96.93 \\
0.90 & 0.7 & 95.82 \\
0.75 & 0.3 & 96.85 \\
\textbf{0.75} & \textbf{0.5} & \textbf{97.37} \\
\bottomrule
\end{tabular}
\end{table}

\subsection{Performance under Limited Labels}
To assess performance when annotations are scarce, we vary the labeled fraction on BoT-IoT while retaining the 90\% training setting. Table~\ref{tab:label_ratio_bot} reports accuracy, TPR, and TNR with 1\%, 5\%, 10\%, and 20\% labels.

\begin{table}[pos=htbp]
\centering
\caption{Effect of Labeling Ratio on BoT-IoT Detection Performance (90\% Training Data)}
\label{tab:label_ratio_bot}
\scriptsize
\begin{tabular}{c|ccc}
\toprule
\textbf{Labeling Ratio} & \textbf{Accuracy} & \textbf{TPR} & \textbf{TNR} \\
\midrule
1\%   & 0.926 & 0.926 & 0.933 \\
5\%   & 0.944 & 0.944 & 0.945 \\
10\%  & 0.957 & 0.957 & 0.956 \\
20\%  &  \textbf{0.974} & \textbf{0.973} & \textbf{0.974} \\
\bottomrule
\end{tabular}
\end{table}


AnchorMixGAN retains 92.6\% accuracy with only 1\% labels, together with a TPR of 92.6\% and a TNR of 93.3\% (Table~\ref{tab:label_ratio_bot}). Increasing the labeled fraction to 5\%, 10\%, and 20\% raises accuracy to 94.4\%, 95.7\%, and 97.4\%, respectively. More annotations consistently improve both attack detection and non-DDoS recognition, with TPR and TNR reaching 97.3\% and 97.4\% at the highest label fraction. Reducing the labeled fraction from 20\% to 1\% lowers accuracy by 4.8 percentage points, while all three metrics remain above 92\%.

\subsection{Inference Efficiency}
\label{subsec:efficiency}
Both methods deploy one WideResNet detector. Table~\ref{tab:new-efficiency} reports parameter count, peak allocated GPU memory, and forward time on the RTX 5070 Ti Laptop GPU. Preprocessed test features are preloaded on the GPU; after 20 warm-up batches, each model processes the complete test set three times at batch size 64 with synchronization. We take the median time for each model and average over seeds. Timing covers detector computation and excludes preprocessing, transfer, generation, and reference-classifier prediction.
\begin{table}[pos=!htbp]
\centering\footnotesize
\caption{Inference efficiency comparison on the RTX 5070 Ti Laptop GPU.}
\label{tab:new-efficiency}
\setlength{\tabcolsep}{2pt}
\renewcommand{\arraystretch}{1.08}
\begin{adjustbox}{max width=\linewidth}
\begin{tabular}{lrrrrr}
\toprule
\multirow{2}{*}{Dataset} & \multirow{2}{*}{Flows} & \multirow{2}{*}{Memory (MB)} & \multirow{2}{*}{Parameters} & \multicolumn{2}{c}{Time (s)}\\
\cmidrule(lr){5-6}
 &  &  &  & MixGAN & AnchorMixGAN\\
\midrule
NSL-KDD & 4000 & 73.43 & 498514 & 0.109 & \textbf{0.103}\\
BoT-IoT & 20000 & 74.21 & 498514 & \textbf{0.565} & 0.582\\
CICIoT2023 & 20000 & 74.39 & 498514 & 0.545 & \textbf{0.531}\\

\bottomrule
\end{tabular}
\end{adjustbox}
\end{table}

AnchorMixGAN and MixGAN have the same detector size: both contain 498,514 parameters, and peak allocated GPU memory ranges from 73.43 to 74.39~MB across the datasets. AnchorMixGAN processes 4,000 NSL-KDD flows in 0.103~s and 20,000 BoT-IoT and CICIoT2023 flows in 0.582~s and 0.531~s, respectively. The corresponding MixGAN times are 0.109, 0.565, and 0.545~s. These closely matched runtimes show that the additional training operations preserve the computational cost of the deployed detector. Donor-view construction and target estimation are completed during training, leaving a single detector to classify incoming flows.

\FloatBarrier

\section{Conclusion}
\label{sec:conclusion}
This paper presented AnchorMixGAN, a generative semi-supervised framework for DDoS detection in cloud-integrated IoT networks with limited labeled traffic. Its Anchor-MAS module constructs field-group views of each real unlabeled flow using generated donor values. Predictions on the original flow and its views are averaged and sharpened to form a soft target, which remains paired with that flow during MixUp. The detector learns from both labeled records and mixed examples. The theoretical analysis relates target error to reference-classifier error, view-induced changes, and sharpening, and bounds the resulting change in cross-entropy at a fixed detector prediction.

Experiments on NSL-KDD, BoT-IoT, and CICIoT2023 demonstrate the effectiveness of this approach. At the 90\% training setting, AnchorMixGAN achieves accuracies of 97.3\%, 97.4\%, and 96.5\%, respectively, outperforming the compared methods in overall accuracy. It also maintains the highest accuracy across the evaluated training ratios on NSL-KDD and BoT-IoT, and reaches 92.6\% accuracy on BoT-IoT with only 1\% labeled data. The component and parameter studies support combining generative augmentation with anchor-aligned semi-supervised training and show the importance of selecting suitable mixing and sharpening parameters.

AnchorMixGAN performs view construction and target estimation during training, requiring only a single WideResNet detector at inference.

\section*{Acknowledgments}
This work was supported in part by the National Natural Science Foundation of China (NSFC) under Grants 62162057 and 61872254, by the Central Government Funds for Supporting the Reform and Development of Local Colleges and Universities under Grant 62162057, by the Key Laboratory of Information Network Security under Grant C20606, by the Key Laboratory of Data Protection and Intelligent Management under Grant SCUSAKFKT202402Y, and by the Open Program of Shaanxi Key Laboratory of Intelligent Policing, Shaanxi Police College, under Grant SXZJ25KF08.

\begingroup\small\raggedright
\bibliographystyle{elsarticle-num}
\bibliography{mybibfile}
\endgroup
\end{document}